\documentclass[letterpaper, 10 pt, conference]{ieeeconf}  % Comment this line out if you need a4paper

\IEEEoverridecommandlockouts                              % This command is only needed if 
\usepackage{graphics} % for pdf, bitmapped graphics files
\usepackage{epsfig} % for postscript graphics files
\usepackage{times} % assumes new font selection scheme installed
\usepackage{amsmath} % assumes amsmath package installed
\usepackage{amssymb}  % assumes amsmath package installed
\usepackage{algorithm}
\usepackage{algpseudocode}
\usepackage{multirow}
\usepackage{booktabs}
\usepackage{makecell}
\usepackage{pifont} % 必须：提供匹配的钩和叉
\usepackage{placeins}
\usepackage{hyperref}
\usepackage{cleveref}
\newcommand{\cmark}{\ding{51}} % 钩
\newcommand{\xmark}{\ding{55}} % 叉

\title{\LARGE \bf
VGP-Nav: Metric-Aware Visual Geometric Perception for Robot Navigation 
}

\author{Hewei Pan$^{1*}$,
        Weiye Zhu$^{1*}$, 
        Zekai Zhang$^{1}$, 
        Zitong Huang$^{1}$,
        Rongtao Xu$^{2}$,
        Jinbao Wang$^{3,\dagger}$,
        Feng Zheng$^{1,4,\dagger}$
\thanks{$^*$Equal Contribution.}
\thanks{$^\dagger$ Corresponding authors.}
\thanks{$^1$Hewei Pan, Weiye Zhu, Zekai Zhang, Zitong Huang and Feng Zheng are with Southern University of Science and Technology, Shenzhen, China.}
\thanks{$^2$Rongtao Xu is with MBZUAI, Abu Dhabi, UAE.}
\thanks{$^3$Jinbao Wang is with Shenzhen University, Shenzhen, China.}
\thanks{$^4$Feng Zheng is also with SpatialTemporal AI, Shenzhen, China.}
}

\begin{document}

\maketitle
\thispagestyle{empty}
\pagestyle{empty}

%%%%%%%%%%%%%%%%%%%%%%%%%%%%%%%%%%%%%%%%%%%%%%%%%%%%%%%%%%%%%%%%%%%%%%%%%%%%%%%%

%%%%%%%%%%%%%%%%%%%%%%%%%%%%%%%%%%%%%%%%%%%%%%%%%%%%%%%%%%%%%%%%%%%%%%%%%%%%%%%%
\begin{abstract}
Reliable robotic navigation necessitates the seamless integration of accurate global localization and dense, metric-consistent obstacle perception. A common strategy to achieve these capabilities involves integrating diverse sensing modalities: cameras offer rich visual features for localization, while active sensors like LiDAR provide direct metric measurements. However, such multi-sensor configurations necessitate complex spatial-temporal calibration and increase deployment overhead.
Although vision-only approaches offer a low-cost and scalable alternative, existing monocular visual systems typically struggle to simultaneously achieve efficient, globally consistent localization and dense, metric-consistent geometric perception. To bridge this gap, we propose \textbf{VGP-Nav}, a unified framework for \textit{Metric-Aware Visual Geometric Perception} that relies solely on monocular RGB input to jointly support metric localization and obstacle perception. Our key insight is to anchor localization-grounded visual geometry to physically meaningful scale constraints derived from ground-plane geometry, thereby providing a reliable metric reference for monocular perception. VGP-Nav resolves monocular scale ambiguity online and produces localization-grounded, metric obstacle representations that are directly applicable to downstream planning. Extensive experiments demonstrate strong generalization across diverse environments and successful deployment on real mobile robots, highlighting the practicality of our approach for scalable, low-cost, and safe autonomous navigation.

\end{abstract}

\section{INTRODUCTION}

Humans navigating a familiar environment rely predominantly on vision to answer two fundamental questions: ``where am I?'' and ``how can I move safely?'' Even when the scene changes, such as moved furniture or newly appeared obstacles, we can simultaneously localize ourselves and perceive surrounding geometry to adapt our motion. Inspired by this capability, an autonomous robotic system ideally would integrate localization and geometric perception within a unified visual framework, enabling scalable and cost-effective navigation.

However, many robotic navigation systems usually adopt a multi-sensor paradigm~\cite{zuo2019lic,zhang2025towards}, combining cameras with active sensors such as LiDAR. Cameras provide rich visual features that can be used for location~\cite{kendall2015posenet,sarlin2019coarse,sidorov2025gsplatloc}, while LiDAR offers direct metric measurements to perceive obstacles and the surrounding geometry. While generally effective, these multi-sensor configurations introduce additional calibration, synchronization, and hardware complexity. These challenges are further amplified on highly dynamic platforms such as humanoid robots, where mechanical vibrations and torso oscillations complicate cross-sensor alignment. In such scenarios, vision-only pipelines using a single camera provide a conceptually streamlined alternative. However, even these vision-only navigation systems often remain architecturally decoupled, handling global localization and metric geometric perception with separate models. This separation limits the ability to reason jointly about pose and geometry, motivating fully unified vision-only frameworks that integrate localization and dense, metric-aware geometric perception.

% However, in practice, many robotic navigation systems decouple these two essential functions~\cite{qu2021outline}. While vision-based algorithms have leveraged the rich feature information within images to make remarkable progress in global localization~\cite{kendall2015posenet,sarlin2019coarse,sidorov2025gsplatloc}, they typically focus solely on estimating camera poses, which cannot inherently provide the metric-consistent, dense geometric awareness required for obstacle avoidance. Consequently, robots often need to rely on additional active sensors, such as lidar, to perceive the surrounding metric geometry. This decoupled paradigm is particularly brittle for humanoid robots: the inherent mechanical vibrations and torso oscillations during bipedal locomotion impose extreme challenges on the spatial-temporal alignment between camera and lidar. Frequent calibration drifts and synchronization errors not only increase hardware costs but also severely compromise navigation reliability in large-scale deployments. In contrast, a unified vision-only approach offers a fundamental advantage: by deriving both location and geometry from a single stream of pixels, it ensures inherent physical consistency and remains immune to the cross-sensor alignment issues that plague multi-sensor systems.

\begin{figure}[tp]
    \centering
    \includegraphics[width=\columnwidth]{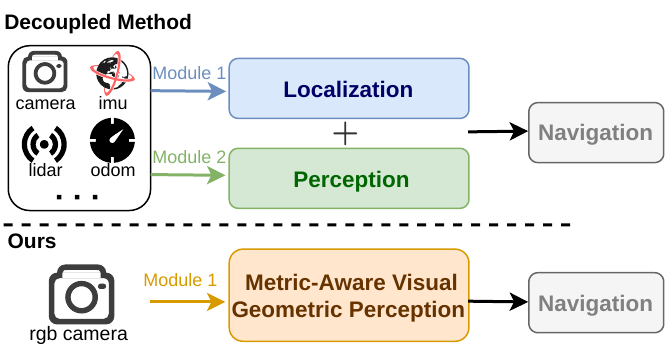} 
    \caption{\textbf{Conceptual comparison between the conventional decoupled navigation pipeline and our proposed framework.} Top: Decoupled navigation pipelines typically separate localization from perception. Bottom: Our framework unifies localization and metric perception into a single module, enabling robust navigation using only a monocular RGB camera.}
    \label{fig:fig1} 
\end{figure}

Recently, feed-forward 3D reconstruction techniques~\cite{wang2024dust3r,wang2025vggt,wang2025pi,wang2025continuous} have shown great potential toward a unified visual solution for both localization and obstacle perception. Trained on large-scale multi-view data, these models can efficiently predict multi-view camera poses and dense 3D geometry in a single forward pass. By jointly inferring pose and structure within a shared representation, they overcome the limitations of architecturally decoupled pipelines, enabling coherent reasoning about camera motion and scene geometry. However, applying such models to robot navigation faces a core technical barrier: \textbf{Scale Ambiguity}. Unlike stereo or lidar systems, the output of monocular reconstruction lacks a physical metric scale, making it directly unusable for the precise distance measurements required by robot controllers.

To address these challenges, we propose \textbf{VGP-Nav}, a unified framework for \textit{Metric-Aware Visual Geometric Perception}. Our key insight is to anchor localization-grounded visual geometry to physically meaningful scale constraints derived from ground-plane geometry. Specifically, we leverage geometry-aware retrieval strategy and weighted motion averaging to generate globally consistent pose priors. These priors serve as the bridge to project and anchor feed-forward visual reconstructions onto the ground plane, thereby providing a reliable metric reference for monocular perception. This design resolves the inherent scale ambiguity and enables the construction of metric-accurate occupancy maps for downstream navigation.

Our main contributions are summarized as follows:
\begin{itemize}

\item We propose \textbf{VGP-Nav}, a unified framework for \textit{Metric-Aware Visual Geometric Perception} that jointly supports metric visual localization and geometric obstacle sensing within a single perception pipeline, enabling metric-accurate navigation solely using a monocular RGB camera.

\item We introduce a tightly coupled localization–perception paradigm that enables robust scale recovery. By anchoring feed-forward visual geometry to the ground-plane structure via global pose priors, our approach effectively resolves monocular scale ambiguity and ensures metric-consistent obstacle reconstruction.

\item We achieve state-of-the-art performance across multiple benchmarks and further validate the robustness of \textbf{VGP-Nav} by successful real-world robotic deployments, indicating its practical readiness for autonomous navigation.

\end{itemize}

\section{Relate Work}
\begin{figure*}[t]
    \centering
    \includegraphics[width=\textwidth]{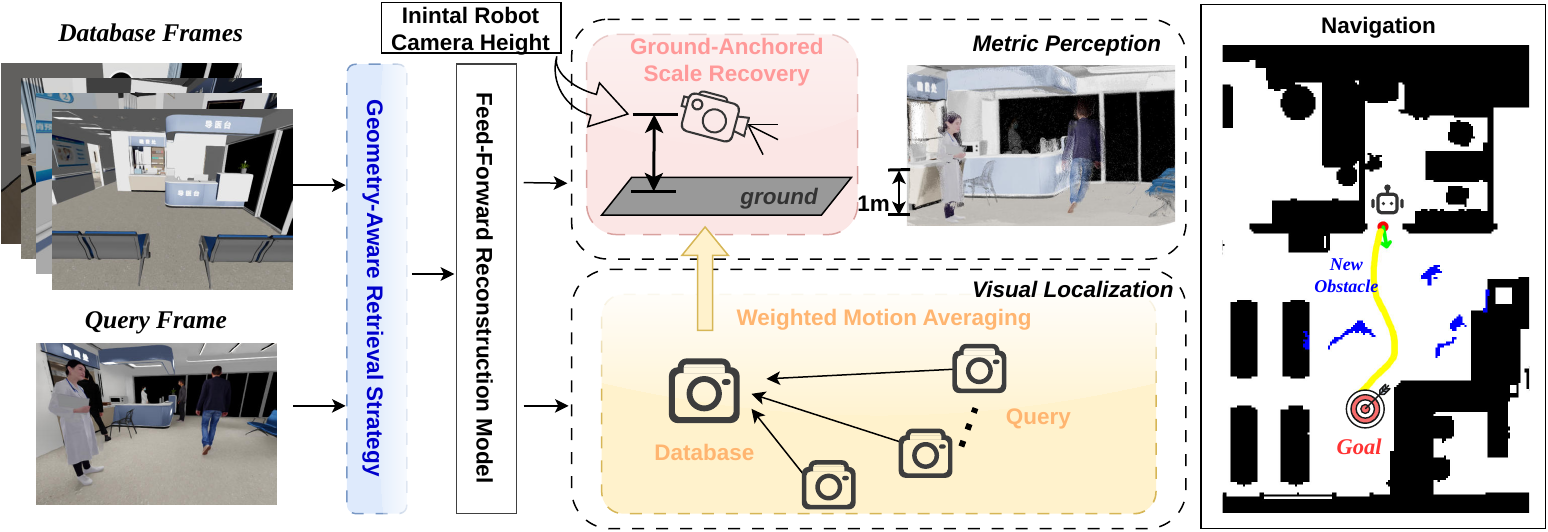} 
    \caption{\textbf{Architecture overview.} The system selects diverse reference images via Geometry-Aware Retrieval Strategy, then generates multi-view constraints using a Feed-Forward Reconstruction backbone. These are processed through Weighted Motion Averaging for 6-DoF global localization, while a Ground-Anchored Scale Recovery module resolves metric scale against the physical ground plane. This unified pipeline enables concurrent global positioning and metric obstacle perception for safe navigation.}
    \label{fig:overview_full} 
\end{figure*}
\subsection{Visual Localization}
\label{vloc_method}
Structure-based visual localization methods rely on pre-built 3D scene representations~\cite{sarlin2019coarse,wang2025igaussian}. 
Traditional pipelines typically establish 2D-3D correspondences by matching query features~\cite{detone2018superpoint,sarlin2020superglue} against a pre-built Structure-from-Motion (SfM) model~\cite{schonberger2016structure}, subsequently solving for the camera pose via Perspective-n-Point (PnP) and RANSAC~\cite{fischler1981random}.
In contrast, learning-based alternatives encode scene information implicitly within neural networks. 
Scene Coordinate Regression (SCR) predicts per-pixel 3D scene coordinates~\cite{brachmann2021visual,brachmann2023accelerated}, 
whereas Absolute Pose Regression (APR) directly regresses the global 6-DoF camera pose~\cite{kendall2015posenet,chen2022dfnet,chen2024map}. 
Despite their different formulations, these approaches are inherently scene-specific, requiring dedicated mapping or training for each environment. 
Such dependency limits cross-scene generalization and scalability, making large-scale deployment costly and inflexible.

In contrast, structureless visual localization avoids explicit 3D scene representations and models the environment as a database of geo-referenced images. Visual Place Recognition (VPR)~\cite{arandjelovic2016netvlad,revaud2019learning} retrieves visually similar reference images to localize the query, and Relative Pose Regression (RPR)~\cite{balntas2018relocnet,arnold2022map,dong2025reloc3r} localizes the query by predicting the relative 6-DoF transformation between image pairs. By removing the need for pre-built 3D models, these methods are scene-agnostic and easy to maintain, but primarily estimate sparse camera poses without recovering dense scene geometry. Building on the flexibility of structureless localization, our method further incorporates dense visual geometry to achieve more accurate and reliable localization.

\subsection{Feed-Forward Reconstruction Models}
Recent advances in Vision Transformers have enabled feed-forward reconstruction models that directly predict camera poses and dense 3D geometry, core perceptual components for safe autonomous navigation. Methods such as Dust3R~\cite{wang2024dust3r} and MASt3R~\cite{leroy2024grounding} recover relative poses and dense structure from image pairs, while more recent models like VGGT~\cite{wang2025vggt} extend this paradigm to true multi-view settings with globally consistent predictions. Despite their impressive geometric reasoning, a fundamental barrier to their robotic deployment is the inherent scale ambiguity. As these models are typically supervised by relative geometric objectives, they recover 3D structures only up to an arbitrary scale, rendering them unsuitable for direct metric navigation.  In this work, we bridge the gap between relative geometry and metric geometric perception, enabling reliable real-world localization and obstacle sensing.

\subsection{Metric Geometric Perception}
Perceiving the dense metric geometry of the environment is fundamental for robot interaction. In the domain of monocular perception, research has evolved from predicting relative depth maps to recovering absolute metric depth. Recent state-of-the-art models like MoGe-2~\cite{wang2025moge} have shown the capability to infer dense metric depth directly from single view by learning from large-scale datasets. Beyond single-view estimation, multi-view reconstruction model such as MapAnything~\cite{keetha2025mapanything} leverage temporal consistency across video sequences to generate coherent metric 3D representations.

While these advancements provide powerful tools for dense geometry perception, they are typically designed as standalone perception tasks, decoupled from the global localization problem. In contrast, VGP-Nav unifies metric perception with structureless global localization. By anchoring feed-forward geometry to the ground plane, we achieve metric-aware perception that is strictly aligned with the robot's navigation coordinate system.
% \begin{figure*}[t]
%     \centering
%     \includegraphics[width=\textwidth]{fig/overview.drawio.pdf} 
%     \caption{\textbf{Architecture overview.} The system selects diverse reference images via Geometry-Aware Retrieval Strategy, then generates multi-view constraints using a Feed-forward Reconstruction backbone. These are processed through Weighted Motion Averaging for 6-DoF global localization, while a Ground-Anchored Scale Recovery module resolves metric scale against the physical ground plane. This unified pipeline enables concurrent global positioning and metric obstacle perception for safe navigation.}
%     \label{fig:overview_full} 
% \end{figure*}
\section{METHODOLOGY}
\subsection{Problem Formulation}
We consider a robot navigating in a known environment represented by a structureless database $\mathcal{D} = \{ (I_i, \mathbf{T}_{wi}) \}_{i=1}^{M}$. Our objective is to achieve Metric-Aware Visual Geometric Perception, providing the spatial understanding required for autonomous motion.

Given a monocular query image $I_q$ at time $t$, VGP-Nav aims to solve for a navigation-ready state $\mathcal{S}_t = \{ \mathbf{T}_{wc}, \mathcal{O}_w \}$, which satisfies:
\begin{itemize}
\item \textbf{Global Pose Alignment ($\mathbf{T}_{wc} \in SE(3)$):} Accurately aligning the camera $C$ of query frame to the world frame $W$ to determine the robot's location.
\item \textbf{Metric Obstacle Representation ($\mathcal{O}_w \subset \mathbb{R}^3$):} Producing a dense, metric-consistent geometric point map of the query frame with absolute scale, enabling collision-free planning.
\end{itemize}

\subsection{Overall Architecture} 
As illustrated in Fig.~\ref{fig:overview_full}, VGP-Nav is a unified framework that orchestrates concurrent visual localization and metric obstacle sensing for robot navigation. The system builds on an efficient feed-forward reconstruction backbone augmented with explicit geometric constraints. Given a query image, the pipeline first retrieves relevant images from the database using a \textbf{Geometry-Aware Retrieval Strategy}. These images are processed by the feed-forward reconstruction backbone to yield relative poses and dense geometry. To ensure metric accuracy for navigation, we implement a dual-constraint refinement process: a \textbf{Weighted Motion Averaging} module aggregates the cameras of multi-view constraints to robustly estimate the global 6-DoF pose of the query image, while a \textbf{Ground-Anchored Scale Recovery} mechanism resolves the metric scale by aligning the reconstructed geometry with the global ground plane. This mechanism utilizes a calibrated height prior to initialize the ground reference and continuously enforces consistency between the perceived and actual ground height by optimizing the camera extrinsics. As a result, this design enables the scale-ambiguous point cloud to be instantaneously projected into a metric occupancy representation. This allows the robot to concurrently determine ``where I am'' and ``how to move safely''.

\begin{figure}
    \centering
    \includegraphics[width=\columnwidth]{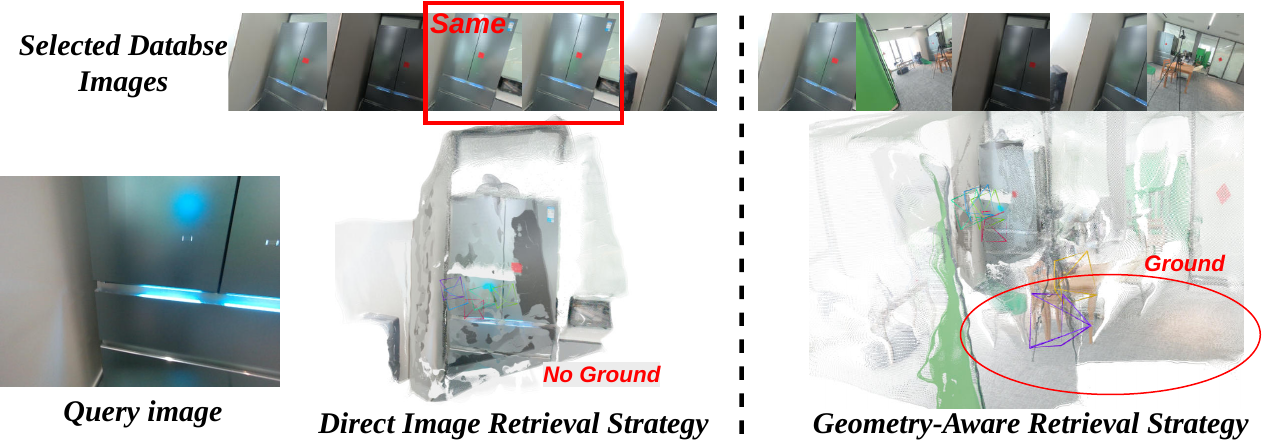}
    \caption{\textbf{Qualitative comparison of point cloud reconstruction under different retrieval strategies.} 
    \textbf{Left (Direct Retrieval):} Clustered poses result in a redundant and restricted field-of-view, often failing to capture the ground plane. \textbf{Right (Ours):} Our strategy ensures a diverse and expansive camera distribution, securing sufficient ground observations to anchor the reconstruction in metric space.}
    \label{fig:retrieval}
\end{figure}

\subsection{Geometry-Aware Retrieval Strategy}
In hierarchical visual localization pipelines~\cite{sarlin2019coarse,dong2025reloc3r}, the coarse localization stage typically selects the top-$k$ reference images directly based on visual similarity scores from a VPR module. However, visually similar images in large-scale databases often exhibit substantial overlap, corresponding to clustered camera poses or even near-identical duplicates. Directly using such a spatially redundant set results in a collapsed field-of-view that lacks sufficient viewpoint diversity and effective geometric baselines. This not only degrades the precision of downstream pose estimation but also limits the spatial extent of the reconstructed geometry, frequently failing to capture critical ground-plane structures, the essential metric anchors for scale recovery (Fig.~\ref{fig:retrieval}). To address this issue, we propose a Geometry-Aware Retrieval Strategy (see Algorithm~\ref{alg:geo_retrieval}).

The process operates in two stages:
\begin{enumerate}
    \item Spatial Outlier Rejection: We first retrieve a larger pool of candidate reference images by selecting the top-$N$ matches, with $N$ greater than $k$, and compute the median of their camera translations. Candidates that deviate significantly from this median are discarded as likely perceptual aliasing errors, while the best-matching candidate is explicitly retained. 

    \item Diversity-Driven Selection: From the valid candidates, we select a subset of $k$ images using a greedy strategy inspired by Farthest Point Sampling. We iteratively select the candidate that maximizes a combined score of visual similarity and spatial distance (considering both translation and viewing direction) relative to the already selected set.  
\end{enumerate}
Overall, this geometry-aware retrieval strategy ensures a well-conditioned reference set by balancing visual similarity with viewpoint diversity. This not only improves localization accuracy but also maximizes environmental coverage, providing the critical ground observations necessary for stable, metric-consistent reconstruction.

\begin{algorithm}[htbp]
\caption{\textbf{Geometry-Aware Retrieval Strategy}}
\label{alg:geo_retrieval}
\begin{algorithmic}[1]
\Require 
Query image $I_q$; Database $\mathcal{D} = \{(I_i, T_i)\}$; 
Target number $k$; Search pool size $N$ ($N = 4k$);
Weights $\lambda_{pos}=0.2, \lambda_{ang}=0.6, \lambda_{sim}=0.2$.
\Ensure Selected reference set $\mathcal{S}$.

\State \textbf{Phase 1: Spatial Outlier Rejection}
\State Retrieve top-$N$ candidates $\mathcal{C}_{init}$ based on VPR similarity $S(I_q, I_i)$.
\State $I_{best} \leftarrow \arg\max_{I \in \mathcal{C}_{init}} S(I_q, I)$.
\State Compute geometric median pose $\bar{T}$ of $\mathcal{C}_{init}$.
\State Compute distances $d_i = \| \mathbf{t}_i - \mathbf{t}_{\bar{T}} \|_2$ for all $i \in \mathcal{C}_{init}$.
\State Threshold $\tau \leftarrow \text{Percentile}(\{d_i\}, 95\%)$.
\State $\mathcal{C}_{valid} \leftarrow \{ I_i \in \mathcal{C}_{init} \mid d_i \leq \tau \} \cup \{ I_{best} \}$.

\State \textbf{Phase 2: Diversity-Driven Selection}
\State Initialize selected set $\mathcal{S} \leftarrow \{ I_{best} \}$.
\State Normalize VPR scores in $\mathcal{C}_{valid}$ to $[0, 1]$.
\While{$|\mathcal{S}| < k$ and $|\mathcal{S}| < |\mathcal{C}_{valid}|$}
    \State $c^* \leftarrow \text{None}, \quad \text{Score}_{max} \leftarrow -\infty$
    \For{each candidate $c \in \mathcal{C}_{valid} \setminus \mathcal{S}$}
        \State Compute min-distance to selected set:
        \State $\delta_{cs} \gets \lambda_{pos} \frac{\| \mathbf{t}_c - \mathbf{t}_s \|}{L_{scale}} + \lambda_{ang} \frac{\theta( \mathbf{v}_c, \mathbf{v}_s )}{\pi}$
\State $D_{geo}(c, \mathcal{S}) = \min_{s \in \mathcal{S}} \delta_{cs}$
        % \State $D_{geo}(c, \mathcal{S}) = \min_{s \in \mathcal{S}} \left( \lambda_{pos} \frac{\| \mathbf{t}_c - \mathbf{t}_s \|}{L_{scale}} + \lambda_{ang} \frac{\theta( \mathbf{v}_c, \mathbf{v}_s )}{\pi} \right)$
        \State Compute combined score:
        \State $J(c) = D_{geo}(c, \mathcal{S}) + \lambda_{sim} \cdot S_{norm}(c)$
        \If{$J(c) > \text{Score}_{max}$}
            \State $\text{Score}_{max} \leftarrow J(c), \quad c^* \leftarrow c$
        \EndIf
    \EndFor
    \State $\mathcal{S} \leftarrow \mathcal{S} \cup \{ c^* \}$
\EndWhile
\State \Return $\mathcal{S}$
\end{algorithmic}
\end{algorithm}

\subsection{Weighted Motion Averaging}
To robustly aggregate the $k$ reference poses $\{ \mathbf{T}_{wi} \}$ and predicted relative poses $\{ \mathbf{T}_{ic} \}$ from a feed-forward reconstruction model, we design a two-stage weighted motion averaging strategy that decouples the estimation of rotation and translation.

\paragraph{Rotation Averaging with Consistency Modeling}
We first compute absolute rotation hypotheses $\{ \mathbf{R}_{wc}^{(i)} = \mathbf{R}_{wi} \mathbf{R}_{iq} \}$, representing them in quaternion space. To mitigate the influence of erroneous estimates, we identify a dominant consensus based on mutual angular consistency. Hypotheses deviating significantly from this consensus are suppressed, and the final global rotation $\mathbf{R}_{wc}^*$ is obtained via weighted averaging. The confidence weight $w_{rot}^{(i)}$ is a monotonically decreasing function of the angular deviation from the consensus, reflecting the reliability of each rotational estimate.

\paragraph{Translation Averaging via Multi-Factor IRLS} 
% Conditioned on $\mathbf{R}_{wc}^*$, the camera center $\mathbf{t}_{wc}$ is recovered by solving a robust multi-view triangulation problem. Each reference pose provides a ray constraint in the world frame. We formulate translation estimation as an Iteratively Reweighted Least Squares (IRLS)~\cite{chatterjee2013efficient} optimization that minimizes the weighted orthogonal distance from the camera center to these rays:
Given the condition of $\mathbf{R}_{wc}^*$, we recover the camera center $\mathbf{t}_{wc}$ by addressing a robust multi-view triangulation problem. Each reference pose provides a ray constraint in the world frame. We approach the translation estimation as an Iteratively Reweighted Least Squares (IRLS)~\cite{chatterjee2013efficient} optimization, which aims to minimize the weighted orthogonal distance from the camera center to these rays.
\begin{equation}
\mathbf{t}_{wc}^* = \arg\min_{\mathbf{t}} \sum_{i=1}^k \Omega_i \cdot \| (\mathbf{I} - \mathbf{v}_i \mathbf{v}_i^T)(\mathbf{t} - \mathbf{c}_i) \|^2,
\end{equation}
where $\mathbf{c}_i$ is the reference center and $\mathbf{v}_i$ is the viewing direction. The composite weight $\Omega_i = w_{rot}^{(i)} \cdot w_{geo}^{(i)} \cdot w_{res}^{(i)}$ integrates three complementary priors:
\begin{itemize}
    \item \textbf{Rotation Consistency ($w_{rot}$):}  Propagates the reliability score from the rotation stage by assigning weights through a Gaussian kernel based on angular deviations, ensuring that rotationally accurate hypotheses guide the translation 
     \item \textbf{Geometric Conditioning ($w_{geo}$):} Optimizes the triangulation configuration by weighting each ray based on the mean sine of its angular separation from other rays. This explicitly promotes large-baseline and non-parallel intersections, thereby minimizing geometric uncertainty in the reconstructed query position.
    \item \textbf{Residual Robustness ($w_{res}$):} Adaptively downweights spatial outliers via robust M-estimation during IRLS iterations.
\end{itemize}
By jointly enforcing rotational coherence and geometric robustness, this process yields a globally consistent query pose $\mathbf{T}_{wc}^*$ that is resilient to individual estimation errors.

\subsection{Ground-Anchored Scale Recovery}

In navigation-centric environments, the ground plane serves as a static and globally consistent geometric reference that remains invariant to camera jitter induced by robotic motion or varying viewpoints across different camera configurations. This inherent stability provides a natural metric anchor for resolving the scale ambiguity in monocular visual reconstructions. We exploit this constraint to recover the metric scale by anchoring the reconstructed geometry to the ground plane in the world coordinate system.

To minimize geometric overhead and accelerate inference, we represent the scene in a gravity-aligned world frame, where the ground plane is inherently orthogonal to the gravity vector. By applying the globally optimized transfer $\mathbf{T}_{wc}^*$, the reconstructed point cloud $\hat{\mathcal{P}}$ is canonicalized into this coordinate system, effectively aligning the scene's vertical axis with gravity. This alignment, coupled with the broad spatial extent secured by our geometry-aware retrieval, transforms ground estimation into a highly efficient statistical problem. Rather than resorting to expensive per-frame plane fitting, we exploit the dense, wide-context ground observations to resolve the ground height via vertical point distribution analysis. Specifically, the ground plane manifests as the dominant peak in the height statistics along the gravity axis(lower than camera height), providing a robust metric anchor with minimal computational cost.

\paragraph{Ground Height Initialization}
At the start of navigation, the robot remains stationary and estimates the ground height in the gravity-aligned world frame. This predicted ground height is aligned to the true physical ground using a known geometric prior: the fixed distance between the camera and the robot base, obtained from calibration. This one-time initialization anchors the reconstructed ground plane to its metric height and determines the global ground height.

\paragraph{Online Scale Maintenance}
During navigation, the recovered scale is maintained by continuously aligning the reconstructed geometry to the anchored ground height. As long as the robot operates on the same dominant support surface, the metric scale remains stable without additional optimization or refinement.
With the metric scale recovered, geometric thresholds become physically meaningful. Points of the query frame in $\mathcal{P}_{w}$ are categorized according to their height relative to the anchored ground plane. Points close to the ground are labeled as traversable, while points between $0.15$\,m above the ground and the camera height are projected onto the ground plane to form a 2D obstacle occupancy map. This direct conversion from visual reconstruction to metric navigation primitives allows the planner to generate collision-free trajectories.

\section{EXPERIMENTS}
\subsection{Overview}
% \subsubsection{Settings}
\begin{table*}[t!]
\centering
\caption{
Visual localization results on the 7-Scenes dataset~\cite{shotton2013scene}. We report the median translation error ($m$) and rotation error ($^\circ$) for each scene. The best result within each category is highlighted in bold.
}
\resizebox{0.9\textwidth}{!}{

\small

\begin{tabular}{c|l|ccccccc|c|c}
\toprule
& \multirow{2}{*}{\textbf{Methods}} & \multirow{2}{*}{\textbf{Chess}} & \multirow{2}{*}{\textbf{Fire}} & \multirow{2}{*}{\textbf{Heads}} & \multirow{2}{*}{\textbf{Office}} & \multirow{2}{*}{\textbf{Pumpkin}} & \multirow{2}{*}{\textbf{RedKitchen}} & \multirow{2}{*}{\textbf{Stairs}} & \multirow{2}{*}{\textbf{Average}} & \textbf{Dataset-specific} \\
&  &  &  &  &  &  &  &  &  & \textbf{training time} \\

\toprule

\multirow{3}{*}{\rotatebox{90}{APR}}

& PMNet~\cite{lin2024learning} & 0.03 / 1.26 & 0.04 / 1.76 & \textbf{0.02} / 1.68 & 0.06 / 1.69 & 0.07 / 1.96 & 0.08 / 2.23 & 0.11 / 2.97 & 0.06 / 1.93 & Days / scene \\

& DFNet~\cite{chen2022dfnet}+NeFeS~\cite{chen2024neural} & \textbf{0.02 / 0.57} & \textbf{0.02 / 0.74} & \textbf{0.02 / 1.28} & \textbf{0.02 / 0.56} & \textbf{0.02 / 0.55} & \textbf{0.02 / 0.57} & \textbf{0.05 / 1.28} & \textbf{0.02 / 0.79} & Days / scene \\

& Marepo~\cite{chen2024map} & \textbf{0.02} / 1.24 & \textbf{0.02} / 1.39 & \textbf{0.02} / 2.03 & 0.03 / 1.26 & 0.04 / 1.48 & 0.04 / 1.71 & 0.06 / 1.67 & 0.03 / 1.54 & 15min / scene \\

\midrule

\multirow{2}{*}{\rotatebox{90}{SCR}} 

& DSAC*~\cite{brachmann2021visual}  & \textbf{0.019} / 1.11 
                 & \textbf{0.019} / 1.24 
         & 0.011 / 1.82 
         & \textbf{0.026} / 1.18 
         &\textbf{ 0.042} / 1.41   
         & \textbf{0.030} / 1.70 
         & 0.042 / 1.42 
         & \textbf{0.027} / 1.41 & Hours / scene \\

    & ACE~\cite{brachmann2023accelerated}  & \textbf{0.019 / 0.7} 
       & \textbf{0.019 / 0.9 }
       & \textbf{0.009 / 0.6 }
       & 0.027 /\textbf{ 0.8 }
       & \textbf{0.042 / 1.1 }
       & 0.042 / \textbf{1.3 }
       & \textbf{0.039 / 1.1}   
       & 0.028 / \textbf{0.93} & 5min / scene \\
\midrule

\multirow{7}{*}{\rotatebox{90}{RPR}}

& CamNet~\cite{ding2019camnet} & 0.04 / 1.73 & 0.03 / 1.74 & 0.05 / 1.98 & 0.04 / 1.62 & 0.04 / 1.64 & 0.04 / 1.63 & 0.04 / 1.51 & 0.04 / 1.69 & Hours \\

& {Map-free (Match)}~\cite{arnold2022map} & 
0.10 / 2.93 &
0.12 / 4.95 &
0.11 / 5.40 &
0.12 / 3.01 &
0.16 / 3.19 &
0.14 / 3.45 &
0.21 / 4.50 &
0.14 / 3.92 & None \\ 

& Map-free (Regress)~\cite{arnold2022map} & 0.09 / 2.66 &
0.13 / 4.54 &
0.11 / 4.81 &
0.11 / 2.77 &
0.16 / 3.11 &
0.14 / 3.48 &
0.18 / 4.70 &
0.13 / 3.72 & None \\ 

& {ExReNet (SN)}~\cite{winkelbauer2021learning} & 0.06 / 2.15 & 0.09 / 3.20 & 0.04 / 3.30 & 0.07 / 2.17 & 0.11 / 2.65 & 0.09 / 2.57 & 0.33 / 7.34 & 0.11 / 3.34 & None \\

& {ExReNet (SUNCG)}~\cite{winkelbauer2021learning} & 0.05 / 1.63 & 0.07 / 2.54 & 0.03 / 2.71 & 0.06 / 1.75 & 0.07 / 2.04 & 0.07 / 2.10 & 0.19 / 4.87 & 0.08 / 2.52 & None \\

& Reloc3r-512~\cite{dong2025reloc3r} 
& 0.03 / 0.88
& 0.03 / \textbf{0.81}
& \textbf{0.01} / 0.95
& 0.04 / 0.88
& 0.06 / 1.10
& \textbf{0.04} / 1.26
& 0.07 / 1.26
& 0.04 / 1.02
& None \\

& \textbf{VGP-Nav(Ours)} 
& \textbf{0.02 / 0.70}
& \textbf{0.02} / 0.88  
& \textbf{0.01 / 0.73} 
& \textbf{0.03 / 0.73} 
& \textbf{0.04 / 1.05} 
& \textbf{0.04 / 1.20 }  
& \textbf{0.03 / 0.72}  
& \textbf{0.03 / 0.86} & None \\

\bottomrule
\end{tabular}

}
\label{tab:7s}
\end{table*}

\begin{table*}[t!]
\centering
\caption{
Visual localization results on the Cambridge Landmarks Dataset~\cite{kendall2015posenet}. We report the median translation error ($m$) and rotation error ($^\circ$) for each scene. The best result within each category is highlighted in bold.
}
\resizebox{0.9\textwidth}{!}{

\small

\begin{tabular}{c|l|ccccc|c|c|c}
\toprule
& \multirow{2}{*}{\textbf{Methods}} & \multirow{2}{*}{\textbf{GreatCourt}} & \multirow{2}{*}{\textbf{KingsCollege}} & \multirow{2}{*}{\textbf{OldHospital}} & \multirow{2}{*}{\textbf{ShopFacade}} & \multirow{2}{*}{\textbf{StMarysChurch}} & \multirow{2}{*}{\textbf{Average (4)}} & \multirow{2}{*}{\textbf{Average}} & \textbf{Dataset-specific} \\
&  &  &  &  &  &  &  &  & \textbf{training time} \\

\toprule

\multirow{3}{*}{\rotatebox{90}{APR}} 

& LENS~\cite{moreau2022lens} & - & 0.33 / \textbf{0.50} & \textbf{0.44} / 0.90 & 0.27 / 1.60 & 0.53 / 1.60 & 0.39 / 1.20 & - & Days / scene \\

& PMNet~\cite{lin2024learning} & - & \textbf{0.31} / 0.55 & \textbf{0.44 / 0.79} & 0.17 / 0.86 & \textbf{0.31 / 0.96} & \textbf{0.31} / 0.79 & - & Days / scene \\

& DFNet~\cite{chen2022dfnet}+NeFeS~\cite{chen2024neural} & - & 0.37 / 0.54 & 0.52 / 0.88 & \textbf{0.15 / 0.53} & 0.37 / 1.14 & 0.35 / \textbf{0.77} & - & Days / scene \\

\midrule

\multirow{2}{*}{\rotatebox{90}{SCR}} 

& DSAC*~\cite{brachmann2021visual}  & \textbf{0.34 / 0.2}   
         & \textbf{0.18 / 0.3} 
         & \textbf{0.21 / 0.4 }
         & \textbf{0.05 / 0.3 }
         & \textbf{0.15 / 0.6 }
         &\textbf{ 0.14 / 0.4 }
         & \textbf{0.19 / 0.4} & Hours / scene \\

& ACE~\cite{brachmann2023accelerated}  & 0.43 / \textbf{0.2 }
       & 0.28 / 0.4 
       & 0.31 / 0.6 
       & \textbf{0.05 / 0.3} 
       & 0.18 / \textbf{0.6}
       & 0.20 / \textbf{0.4 }
       & 0.25 / \textbf{0.4} & 5min / scene \\

\midrule
\multirow{7}{*}{\rotatebox{90}{RPR}} 
& AnchorNet~\cite{saha2018improved} & - & 0.57 / 0.88 & {1.21 / 2.55} & 0.52 / 2.27 & 1.04 / 2.69 & 0.84 / 2.10 & - & Hours  \\

& {Map-free (Match)}~\cite{arnold2022map} & 9.09 / 5.33 &
2.51 / 3.11 &
3.89 / 6.44 &
1.04 / 3.61 &
3.00 / 6.14 &
2.61 / 4.83 &
3.90 / 4.93 & None \\

& {Map-free (Regress)}~\cite{arnold2022map} & 8.40 / 4.56 &
2.44 / 2.54 &
3.73 / 5.23 &
0.97 / 3.17 &
2.91 / 5.10 &
2.51 / 4.01 & 
3.69 / 4.12 &
None \\

& {ExReNet (SN)}~\cite{winkelbauer2021learning} & 10.97 / 6.52 & 2.48 / 2.92 & 3.47 / 3.90 & 0.90 / 3.27 & 2.60 / 4.98 & 2.36 / 3.77 & 4.08 / 4.32 & None \\

& {ExReNet (SUNCG)}~\cite{winkelbauer2021learning} & 9.79 / 4.46 & 2.33 / 2.48 & 3.54 / 3.49 & 0.72 / 2.41 & 2.30 / 3.72 & 2.22 / 3.03 & 3.74 / 3.31 & None \\

& ImageNet+NCM~\cite{zhou2020learn} & - & - & - & - & - & 0.83 / 1.36 & - & None \\

& {Reloc3r-512}~\cite{dong2025reloc3r}
& 1.22 / 0.73 
&0.42 / 0.36 
&0.62 / 0.55
&\textbf{0.13} / 0.58 
&0.34 / 0.58 
&0.38 / 0.52 
& 0.55 / 0.56 &
None \\

& \textbf{VGP-Nav(Ours)} 
& \textbf{0.62 / 0.49}
&  \textbf{0.40 /0.33} 
&  \textbf{0.50 /0.50} 
& 0.14/ \textbf{  0.26}  
&  \textbf{0.22 /0.43}  
& \textbf{0.31 /0.38}  & \textbf{0.46 / 0.40} & None \\

\bottomrule
\end{tabular}

}

\vspace{-10pt}
\label{tab:cam}
\end{table*}

We evaluate VGP-Nav on three complementary fronts: (1) standard visual localization indoor and outdoor benchmarks to verify pose accuracy; (2) high-fidelity simulation environments to quantify metric perception performance under scene changes; and (3) real-world robot deployment to demonstrate navigation capabilities.

% \subsubsection{Implementation Details}
 VGP-Nav is implemented in PyTorch and evaluated on a workstation equipped with an NVIDIA RTX 5090 GPU. We employ VGGT~\cite{wang2025vggt} as the feed-forward reconstruction backbone. Following the retrieval configuration of Reloc3r~\cite{dong2025reloc3r}, we adopt NetVLAD~\cite{arandjelovic2016netvlad} as the retrieval model, with the number of retrieved database frames set to $k=10$.

\subsection{Visual Localization}
\textbf{Datasets:}
We evaluate VGP-Nav on the widely used 7-Scenes~\cite{shotton2013scene} and Cambridge Landmarks~\cite{kendall2015posenet} datasets. The 7-Scenes dataset consists of seven indoor, room-scale environments, each featuring multiple dense video sequences. The Cambridge Landmarks, which comprises five large-scale urban scenes representing suburban-scale environments. We compare our method against state-of-the-art baselines from the categories discussed in \cref{vloc_method}, including absolute pose regression (APR), scene coordinate regression (SCR) and relative pose regression (RPR).

\textbf{Results:}
We report the median absolute translation error (m) and absolute rotation error ($^{\circ}$). As presented in Table~\ref{tab:7s} and Table~\ref{tab:cam}, VGP-Nav achieves new state-of-the-art performance among all RPR methods in both indoor and outdoor environments, demonstrating the high precision and robustness of our proposed pipeline. Compared to APR, our method achieves comparable accuracy to DFNet~\cite{chen2022dfnet}+NeFeS~\cite{chen2024neural}, which requires several days of training, in indoor environments, and surpasses all APR methods in outdoor scenarios, demonstrating the scalability of our approach.
Compared to SCR, although slightly less accurate, VGP-Nav has a significant advantage in generalization ability. ACE~\cite{brachmann2023accelerated} and DSAC*~\cite{brachmann2021visual} require scene-specific training, while our method requires no training. This allows VGP-Nav to be deployed quickly in different environments.

\subsection{Metric Perception}

\begin{table}[t]
\centering

\caption{Metric Perception Result on Home and Hospital scenes.}
\label{tab:metric_res}
\resizebox{\columnwidth}{!}{
\begin{tabular}{l c l ccc ccc} 
\toprule
\multirow{2}{*}{\textbf{View}} & \multirow{2}{*}{\makecell[c]{\textbf{Known GT} \\ \textbf{Camera}}} & \multirow{2}{*}{\textbf{Method}} & \multicolumn{3}{c}{\textbf{Home}} & \multicolumn{3}{c}{\textbf{Hospital}} \\
\cmidrule(lr){4-6} \cmidrule(lr){7-9} 
& & & Acc.$\downarrow$ & Comp.$\downarrow$ & Overall$\downarrow$ & Acc.$\downarrow$ & Comp.$\downarrow$ & Overall$\downarrow$ \\
\midrule
Single & \cmark & MoGe-2 & 0.18 & 0.15 & 0.16 & 0.15 & 0.13 & 0.14 \\
\midrule
\multirow{2}{*}{Multi} & \cmark & MapAnything & 0.10 & 0.07 & 0.08 & 0.11 & \textbf{0.08} & \textbf{0.09} \\
& \xmark & \textbf{VGP-Nav (Ours)} & \textbf{0.08} & \textbf{0.03} & \textbf{0.05} & \textbf{0.10} & \textbf{0.08} & \textbf{0.09} \\
\bottomrule
\end{tabular}
}
\end{table}
\begin{table}
\centering
% \caption{Per-frame computational time breakdown of the VGP-Nav pipeline.}
\caption{Per-frame computational time breakdown of VGP-Nav.}
\label{tab:realworld}
\resizebox{0.6\columnwidth}{!}{
\begin{tabular}{lc}
\toprule
\textbf{Module} & \textbf{Time (s)}  \\
\midrule
Geometry-Aware Retrieval Strategy & 0.027 \\
VGGT & 0.420 \\
Weighted Motion Averaging & 0.005 \\
Ground-Anchored Scale Recovery & 0.040 \\
\midrule
\textbf{Total System (VGP-Nav)} & 0.492 \\
\bottomrule
\end{tabular}
}
\end{table}
\begin{figure}
    \centering
    \includegraphics[width=0.9\columnwidth]{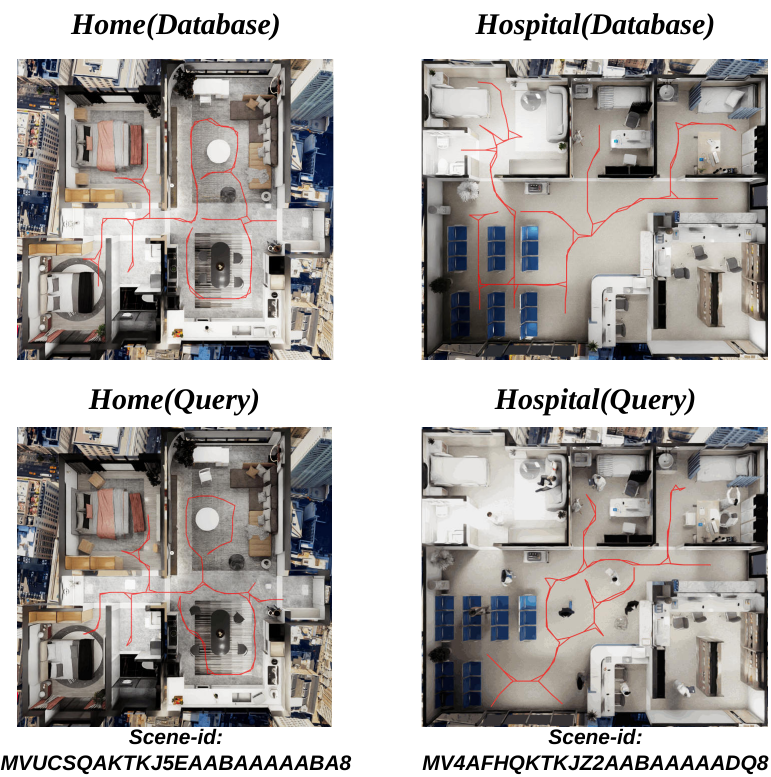}
    \caption{\textbf{Benchmarking metric perception performance under scene changes using automatically generated camera trajectories (red curves).} Left: \textit{Home} scene; Right: \textit{Hospital} scene. Top: Reference database; Bottom: Query runs with modified layouts and new objects.}
    \label{fig:simpath}
\end{figure}

\textbf{Simulation Setup:} We use 2 open-source scenes
from InternScenes~\cite{InternScenes}. To simulate the ``Perception Gap", we generate distinct Database and Query trajectories: (1) \textit{Database Run:} A trajectory in the static base scene; (2) \textit{Query Run:} A trajectory in a modified scene with altered furniture layouts and added new obstacles (e.g., humans). Ground truth trajectories are generated by converting the scene mesh to point clouds, filtering by height to create an occupancy map, and automatically planning a coverage path that traverses the entire environment. As illustrated in Fig.~\ref{fig:simpath}, the synthesized trajectories incorporate significant viewpoint variations to test the robustness of VGP-Nav:

\begin{itemize}
    \item \textbf{Camera Height:} The database camera is positioned at $1.5 \pm 0.1\,\text{m}$, while the query camera is lowered to $1.2 \pm 0.05\,\text{m}$.
    \item \textbf{Camera Orientation:} The pitch angle is constrained within $[15^\circ, 30^\circ]$ downward from the horizontal. This setup mimics the standard downward-looking perspective of mobile robots used for navigation and traversability analysis.
\end{itemize}

\textbf{Results:} We evaluate the quality of the reconstructed metric point cloud $\mathcal{P}_{w}$ using Accuracy (Acc), Completeness (Comp), and Overall scores. We benchmark VGP-Nav against two state-of-the-art metric geometry models: MoGe-2 (single-view)~\cite{wang2025moge}  and MapAnything (multi-view)~\cite{keetha2025mapanything}. As shown in Table~\ref{tab:metric_res}, VGP-Nav achieves superior performance across all metrics in both simulation environments. VGP-Nav significantly outperforms the monocular baseline MoGe-2, demonstrating the necessity of multi-view constraints for scale stability. Notably, while MapAnything relies on full Ground-Truth (GT) camera parameters, VGP-Nav achieves better results with only a coarse estimate of the initial robot camera height above the ground. In the Home scene, our method shows dominant performance, particularly in Completeness (0.03 vs. 0.07). In the Hospital scene, which presents more significant layout variations, the increased localization challenges result in a slight degradation of metric perception. While VGP-Nav still maintains higher Accuracy (0.10) and matches the Overall score of 0.09. These results validate that utilizing the ground plane as a physical anchor effectively mitigates domain-induced scale drift and provides more stable geometric sensing than direct regression, even when the estimated camera poses are suboptimal.

\begin{figure*}
    \centering
    \includegraphics[width=0.9\linewidth]{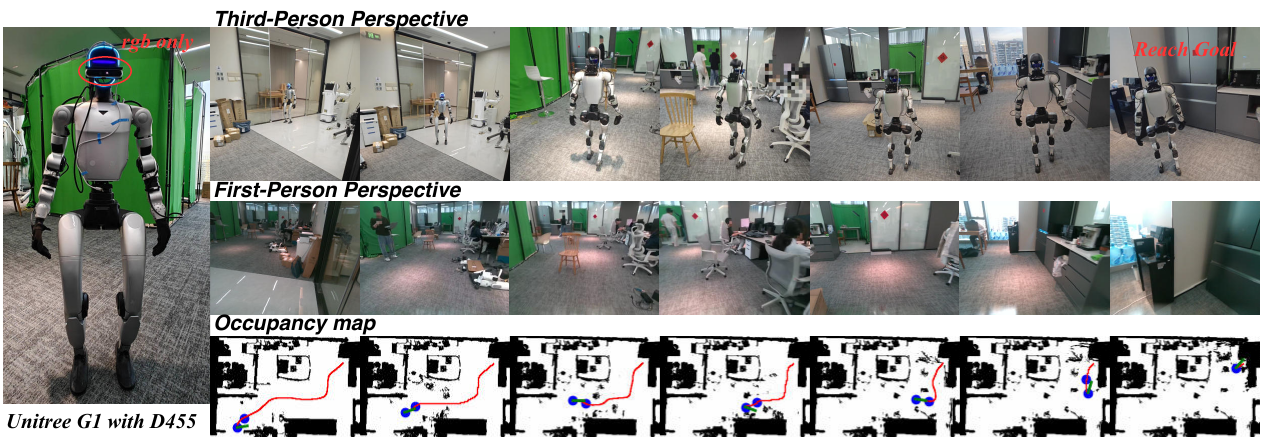}
    \caption{
        \textbf{Real-world experiment.}
        The Unitree G1 performs safe point-goal navigation in the presence of unseen obstacles using only RGB input from an Intel RealSense D455. In occupany map, the current robot state is represented by a blue point with a directional arrow, indicating its 2.5D pose ($x, y, \theta$). The solid blue point denotes the subsequent target waypoint, while the red line illustrates the global trajectory generated by the A$^*$ algorithm~\cite{hart1968formal} on the metric occupancy grid. 
        }
    \label{fig:realworld}
\end{figure*}
\subsection{Real World}
\textbf{Setup:} To validate the practical applicability of VGP-Nav, we deployed the framework on a Unitree G1 humanoid robot equipped with an external Intel RealSense D455 camera (rgb only). The experiments were conducted in a cluttered office environment. Prior to navigation, we utilized a handheld scanner to collect database images, corresponding camera poses, and a coarse occupancy map for global reference. We randomly sampled 20 pairs of start and goal positions to evaluate the autonomous navigation capability of VGP-Nav. The path planner we use A$^*$ algorithm~\cite{hart1968formal}.

\begin{table}
\centering
\caption{Ablation results of our approach on the 7-Scenes dataset for visual localization.}
\label{tab:ablation_vloc}
\resizebox{\columnwidth}{!}{
\begin{tabular}{cc cc} 
\toprule
\makecell[c]{\textbf{Geometry-Aware} \\ \textbf{Retrieval Strategy}} & 
\makecell[c]{\textbf{Weighted Motion} \\ \textbf{Averaging}} & 
\makecell[c]{\textbf{Translation } \\ \textbf{Error(cm) $\downarrow$ }}&
\makecell[c]{\textbf{Rotation }\\ \textbf{ Error  ($^\circ$) $\downarrow$ }} \\
\midrule
 \xmark & \xmark  & 11.3 & 0.90  \\
\cmark & \xmark  & 10.8 & 0.87  \\ 
\xmark & \cmark  & 3.4 & 0.89  \\ 
\cmark & \cmark  & \textbf{3.0}& \textbf{0.86}  \\

\bottomrule
\end{tabular}
}
\end{table}

\begin{table}
\centering
\caption{Ablation results on the Home Scene for Metric Perception.}
\label{tab:ablation_res}
\resizebox{\columnwidth}{!}{
\begin{tabular}{cccc} 
\toprule
\makecell[c]{\textbf{Weighted Motion} \\ \textbf{Averaging}} & 
\makecell[c]{\textbf{Ground-Anchored} \\ \textbf{Scale Recovery}} & 

\makecell[c]{\textbf{Geometry-Aware} \\ \textbf{Retrieval Strategy}} & 
\textbf{Acc.$\downarrow$ / Comp.$\downarrow$} \\
\midrule
\xmark & \xmark & \xmark & 1.74 / 2.68 \\
\cmark & \xmark & \xmark & 0.61 / 1.12 \\ 
\cmark & \cmark & \xmark & 0.09 / 0.05 \\
\cmark & \cmark & \cmark & \textbf{0.08 / 0.03} \\ 
\bottomrule
\end{tabular}
}
\end{table}
\textbf{Quantitative and Qualitative Analysis:}
As summarized in Table~\ref{tab:realworld}, we report the computational time of each module. 
VGP-Nav achieved a success rate of 13/20. The qualitative results in Fig.~\ref{fig:realworld} demonstrate that 
% in open areas, VGP-Nav effectively perceives newly appearing obstacles and guides the robot safely even with significant layout changes. 
benefiting from the synergy between precise visual localization and the stable ground-anchored scale recovery, VGP-Nav provides consistent pose estimates and a stable metric reconstruction of surrounding obstacles. Even during the characteristic torso oscillations and mechanical vibrations of the humanoid gait, our method effectively maintains a steady geometric reference. This allows for the continuous update of the 2D occupancy map within the robot's field of view, successfully guiding the platform to its target while ensuring safety in cluttered office environments.

\textbf{Failure case Analysis:}
We observed that failure cases primarily occurred in narrow passages. During maneuvers to bypass obstacles in close proximity to walls, the camera often directly faced textureless surfaces. This restricted field of view, combined with the scarcity of discriminative visual features, led to a loss of localization and subsequent navigation failure. Future work will focus on integrating temporal consistency to mitigate such cases.

\subsection{Ablation Studies}
We evaluate the impact of our core modules, namely Geometry-Aware Retrieval, Weighted Motion Averaging, and Ground-Anchored Scale Recovery through two sets of ablation experiments.

 \textbf{Impact on Visual Localization:} Table~\ref{tab:ablation_vloc} highlights how our geometric strategies refine pose estimation accuracy. Weighted Motion Averaging is the most critical component for translation accuracy. Comparing the first and third rows, this module reduces the translation error significantly from $11.3\,\text{cm}$ to $3.4\,\text{cm}$. This improvement demonstrates that our motion averaging successfully aligns discrete relative pose estimates into a consistent global coordinate system. Furthermore, the Geometry-Aware Retrieval strategy provides a secondary boost (Row 3 vs. Row 4), achieving the best performance of $3.0\,\text{cm} / 0.86^\circ$. By ensuring that the triangulation process utilizes diverse, wide-baseline views, this strategy further improves the geometric conditioning of the final pose estimation.
 
 \textbf{Impact on Metric Perception:} Table~\ref{tab:ablation_res} demonstrates the necessity of each module for metric perception. Comparing the first and second rows, the significant improvement in Acc./Comp. (from 1.74 / 2.68 to 0.61 / 1.12) underscores that accurate pose estimation is the prerequisite for metric geometry perception. Despite the improved alignment, the reconstruction still lacks absolute scale. The introduction of Ground-Anchored Scale Recovery (Row 3) results in a dramatic error reduction to 0.09 / 0.05. This validates the module's role as the primary mechanism for anchoring the feed-forward reconstruction to the real-world metric space. The full pipeline (Row 4) achieves the highest precision (0.08 / 0.03). Beyond its contribution to localization, the Geometry-Aware Retrieval strategy ensures a more diverse set of viewpoints. As shown in Fig.~\ref{fig:retrieval}, this enhanced diversity provides better observation of the ground plane from various angles, allowing for a more robust scale estimation.

% \FloatBarrier
\section{CONCLUSIONS}

In this work, we present VGP-Nav, a unified monocular framework that bridges global localization and dense metric perception without relying on multi-sensor fusion or complex calibration. By anchoring localization-grounded visual geometry to physically meaningful ground-plane constraints, VGP-Nav effectively resolves monocular scale ambiguity, providing a reliable metric reference for obstacle sensing and downstream planning. Extensive experiments in diverse simulated and real-world environments demonstrate that our approach achieves competitive performance and robust generalization, highlighting its potential for scalable, low-cost autonomous navigation. Future work will explore the integration of temporal consistency and semantic understanding, alongside optimizing system inference speed, to further enhance the stability and real-time efficiency of the navigation pipeline.
% \addtolength{\textheight}{-12cm}   % This command serves to balance the column lengths
                                  % on the last page of the document manually. It shortens
                                  % the textheight of the last page by a suitable amount.
                                  % This command does not take effect until the next page
                                  % so it should come on the page before the last. Make
                                  % sure that you do not shorten the textheight too much.

%%%%%%%%%%%%%%%%%%%%%%%%%%%%%%%%%%%%%%%%%%%%%%%%%%%%%%%%%%%%%%%%%%%%%%%%%%%%%%%%

% \FloatBarrier
\bibliographystyle{bibliography/IEEEtran}
\bibliography{bibliography/ref}

\end{document}